\documentclass[letterpaper, 10 pt, conference]{ieeeconf}  

\IEEEoverridecommandlockouts                              

\usepackage{cite}
\usepackage{amsmath,amssymb,amsfonts}
\usepackage{algorithmic}
\usepackage{graphicx}
\usepackage{textcomp}
\usepackage{xcolor}
\def\BibTeX{{\rm B\kern-.05em{\sc i\kern-.025em b}\kern-.08em
T\kern-.1667em\lower.7ex\hbox{E}\kern-.125emX}}
\usepackage{graphics} 
\usepackage{epsfig} 
\usepackage{mathptmx} 
\usepackage{times} 
\usepackage{amsmath} 
\usepackage{inputenc}
\usepackage{siunitx}
\usepackage{nicefrac}
\usepackage{xcolor}
\usepackage{eqparbox}
\usepackage{tabularx}
\usepackage{csquotes}
\usepackage{url}
\usepackage[caption=false,font=footnotesize]{subfig}
\usepackage{stmaryrd} 
\usepackage{makecell}
\usepackage[acronym, toc, shortcuts, nowarn]{glossaries}

\usepackage{listings}
\usepackage{xcolor}

\definecolor{codegreen}{rgb}{0,0.6,0}
\definecolor{codegray}{rgb}{0.5,0.5,0.5}
\definecolor{codepurple}{rgb}{0.58,0,0.82}
\definecolor{backcolour}{rgb}{0.95,0.95,0.92}

\title{\LARGE \bf
    \acl{SERo}: A generic, backend-agnostic library for running reproducible robotics experiments and tests
}

\author{Frederik Pasch$^{*, 1}$, Florian Mirus$^{*, 1}$, Yongzhou Zhang$^{2, 3}$, Kay-Ulrich Scholl$^{1}$
		\thanks{$^{*}$Equal contribution}%
        \thanks{$^{1}$Intel Labs, 76133 Karlsruhe, Germany}%
		\thanks{$^{2}$Karlsruhe University of Applied Sciences, 76133 Karlsruhe, Germany}%
		\thanks{$^{3}$Karlsruhe Institute of Technology, 76131 Karlsruhe, Germany}
	}

\makeglossaries
\newacronym{CV}{CV}{Computer Vision}
\newacronym{LIDAR}{LIDAR}{Light Detection and Ranging}
\newacronym{AMR}{AMR}{Autonomous Mobile Robot}
\newacronym{IoT}{IoT}{Internet of Things}
\newacronym{ROS2}{ROS2}{Robot Operating System}
\newacronym{ROS}{ROS}{Robot Operating System}
\newacronym{AMCL}{AMCL}{Adaptive Monte Carlo Localization}
\newacronym{SLAM}{SLAM}{Simultaneous Localization and Mapping}
\newacronym{DDS}{DDS}{Data Distribution Service}
\newacronym{RMSE}{RMSE}{Root Mean Square Error}
\newacronym{RTAB-MAP}{RTAB-MAP}{Real-Time Appearance-Based Mapping}
\newacronym{ASAM}{ASAM}{Association for Standardization of Automation and Measurement Systems}
\newacronym{CARLA}{CARLA}{Car Learning to Act}
\newacronym{SERo}{SERo}{Scenario Execution for Robotics}
\newacronym{NGSIM}{NGSIM}{Next Generation Simulation}
\newacronym{ATF}{ATF}{Automated Testing Framework}
\newacronym{BDD}{BDD}{Behavior-Driven Development}
\newacronym{ANTLR4}{ANTLR4}{ANother Tool for Language Recognition}
\newacronym{Open-RMF}{Open-RMF}{Open Robotics Middleware Framework}
\newacronym{gRPC}{gRPC}{Google Remote Procedure Call}
\newacronym{MuJoCo}{MuJoCo}{Multi-Joint dynamics with Contact}

\begin{document}

\maketitle
\thispagestyle{empty}
\pagestyle{empty}

\begin{abstract}

Testing and evaluation of robotics systems is a difficult and oftentimes tedious task due to the systems' complexity and a lack of tools to conduct reproducible robotics experiments. 
Additionally, almost all available tools are either tailored towards a specific application domain, simulator or middleware.
Particularly scenario-based testing, a common practice in the domain of automated driving, is not sufficiently covered in the robotics domain.
In this paper, we propose a novel backend- and middleware-agnostic approach for conducting systematic, reproducible and automatable robotics experiments called \acl{SERo}.
Our approach is implemented as a Python library built on top of the generic scenario description language OpenSCENARIO~2 and Behavior Trees and is made publicly available on GitHub.
In extensive experiments, we demonstrate that our approach supports multiple simulators as backend and can be used as a standalone Python-library or as part of the \acs{ROS2} ecosystem.
Furthermore, we demonstrate how our approach enables testing over ranges of varying values.
Finally, we show how \acl{SERo} allows to move from simulation-based to real-world experiments with minimal adaptations to the scenario description file.

\end{abstract}

\section{INTRODUCTION}
\label{sec:introduction}

Testing and evaluation of robotic systems is a difficult and oftentimes tedious task \cite{Afzal2020}.
Some researchers even argue that one of the major challenges of robotics is to systematize the way its research is conducted \cite{Antonelli2015}.
On the one hand, the complexity of robotic systems is continuously increasing, which makes integration, testing and evaluation of the overall system a challenging task.
On the other hand, there is a lack of a generic and coherent way of conducting robotic evaluations that is applicable to both, simulation and real-world experiments.
Although there is a variety of tools for conducting experiments, these tools are oftentimes customized and tailored towards a specific simulator \cite{CARLAScenarioRunner} or a specific middleware \cite{Kanter2019}.
As a result, most testing and evaluation tools or procedures are even tailored towards a specific use-case, application or domain and remain closed-source.
In the automotive domain, where testing and validation is crucial and obligatory for new components and systems before launch, the \ac{ASAM} created the scenario description language OpenSCENARIO~1, which is the foundation of the \ac{CARLA} \cite{Dosovitskiy2017} scenario runner \cite{CARLAScenarioRunner}.
Additionally, scenario-based testing is a common evaluation step in the research field of automated driving \cite{Riedmaier2020} and several simulators offer this feature \cite{Nalawade2022} for being able to conduct reproducible tests in safety-critical situations.
In the robotics domain, systematic, scenario-based testing is less common, partly due to the engineering complexity and manual effort necessary \cite{Afzal2020}.
According to \cite{Afzal2020}, the most common testing strategies in the robotics domain are \textit{field testing} with the full-system operating in a real-world environment similar to the deployment environment, \textit{logging and playback}, i.e., recording and playing back data (e.g., through rosbag files) collected in the field and \textit{simulation testing}.
Although logging and playing back data collected in the field offers the opportunity to reproducibly test the robotics software with data from erroneous situations and corner cases, it is an open-loop test without the option of validating the robotics system in the loop.
In contrast, simulation and field testing evaluate the complete system in the loop and are essential aspects of robotics testing.
However, transferring test scenarios from simulation to identical or at least similar real-world tests poses significant challenges, both in terms of engineering complexity as well as costs. 

In this paper, we propose a novel approach for conducting systematic and reproducible robotic experiments.
\begin{figure}[t]
    \centering
    \includegraphics[width=1.\linewidth]{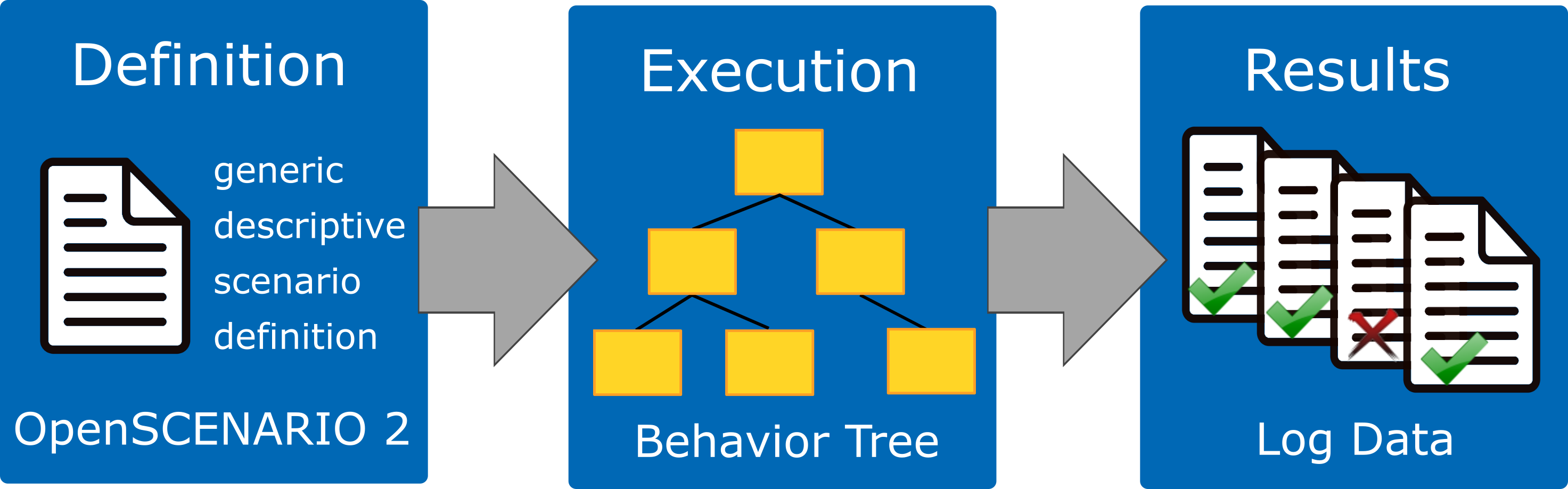}
    \caption{High-level overview of the \acl{SERo} library.}
    \vspace{-0.6cm}
    \label{fig:hl_overview}
\end{figure}
We implement our approach as a Python library and built it on top of two open-source components: the generic scenario description language OpenSCENARIO~2 \cite{OpenScenario} and the Python implementation of Behaviour Trees \cite{Colledanchise2017} PyTrees \cite{Pytrees}.
Fig.~\ref{fig:hl_overview} gives a high-level overview of our system: the user defines a scenario in the OpenSCENARIO~2 language, \acl{SERo} parses the scenario, translates it to a behaviour tree, executes it and finally gathers the test results.
Our implementation is highly modular separating the core components from simulation- and/or middleware-specific modules realized through a plugin-based approach. 
This makes our library extremely flexible and allows the extension to other simulators with reasonable effort.
We demonstrate the usefulness of our approach through various experiments both in simulation and real-world.
We provide the source-code of our library with a small catalogue of example scenarios publicly on GitHub\footnote{\url{https://github.com/IntelLabs/scenario_execution}} to allow the community to reproduce the examples and experiments from this paper.
We are convinced that our library for scenario-based testing will enable robotics researchers and practitioners to greatly simplify and even automate their testing routines and make experimental results more reproducible in the future.











\section{RELATED WORK}
\label{sec:related_work}

Scenario-based testing is a crucially important and active area of research, particularly in the field of automated driving \cite{Fremont2020, Riedmaier2020}. 
Since it is impossible to cover all possible driving situations in real-world tests, there is a plethora of work dedicated to the identification of critical scenarios \cite{Szalay2023}, data-driven generation of scenarios \cite{Cai2022}, strategies for scenario exploration \cite{Schuett2022} to find new and/or relevant scenarios for testing automated driving systems.
Once having identified (a set of) scenarios of interest, it is often useful if not imperative to test particularly safety-critical scenarios in simulation first before conducting field tests with physical robots or automated vehicles. 
In the area of automated driving, there is a large variety of simulators available, that support scenario-based testing \cite{Nalawade2022}.
The choice of the best-suited simulator often depends on the functionality or module to be tested ranging from simulators offering photo-realistic rendering for Computer Vision and machine learning applications \cite{Mueller2018} to tools with advanced and realistic physics simulations for evaluating complex control systems in the loop \cite{Collins2021}.
For instance, the \ac{CARLA} simulator \cite{Dosovitskiy2017} and its scenario-runner \cite{CARLAScenarioRunner}, which is based on OpenSCENARIO~1, are widely used tools for scenario-based testing in the automotive domain in simulation.
However, scenario-based testing in simulation, despite being an important first step, is insufficient to fully test an autonomous system, which will be faced with additional real-world challenges such as sensor noise and other unpredictable factors \cite{Fremont2020}.
\cite{althoff2017commonroad} propose a benchmarking framework called CommonRoads, which combines scenario-based testing with real-world datasets such as the \ac{NGSIM} dataset \cite{NGSIM-US101, He2017} for motion planning in automated driving.

In the robotics domain, scenario-based testing is less common \cite{Afzal2020, Antonelli2015}, partly due to a lack of available tools and software libraries \cite{Kanter2019}.
Hence, most approaches for testing robotics systems in the loop, such as \cite{Huck2022}, where hazardous worker behavior is tested in simulation, remain script-based.
However, script-based approaches are not scalable enough or portable between different simulators let alone simulation and real-world.
The \ac{ATF} \cite{Weisshardt2024} and the TestIt toolkit \cite{Kanter2019} are two examples for libraries that offer testing functionalities and target the \ac{ROS}.
However, both projects only offer support for \ac{ROS} 1 and do not seem to be actively maintained with the last commit being already several years old.
In \cite{Santos2022}, the Python library Pytest-BDD \cite{Pytest-BDD} for \ac{BDD} using a subset of the Gherkin language \cite{Gherkin} is employed to automate acceptance testing for industrial robotic systems. 
Another approach \cite{Suddrey2022} generates Behavior Trees for robotic scenarios from natural language input with the goal of enabling learning from demonstration.

In this paper, we aim to close the gap of a generic tool for scenario-based testing in the robotics domain and propose a backend- and middleware-agnostic library for testing robotics systems both in simulation and with physical robots independent of the application domain.
Although being targeted mainly at the \ac{ROS2} ecosystem, our implementation is modular and flexible enough to be used as a stand-alone Python package as well.


\section{SCENARIO EXECUTION}

\subsection{Software architecture}
\label{subsec:software_arch}

\begin{figure}[t]
    \centering
    \includegraphics[width=1.\linewidth]{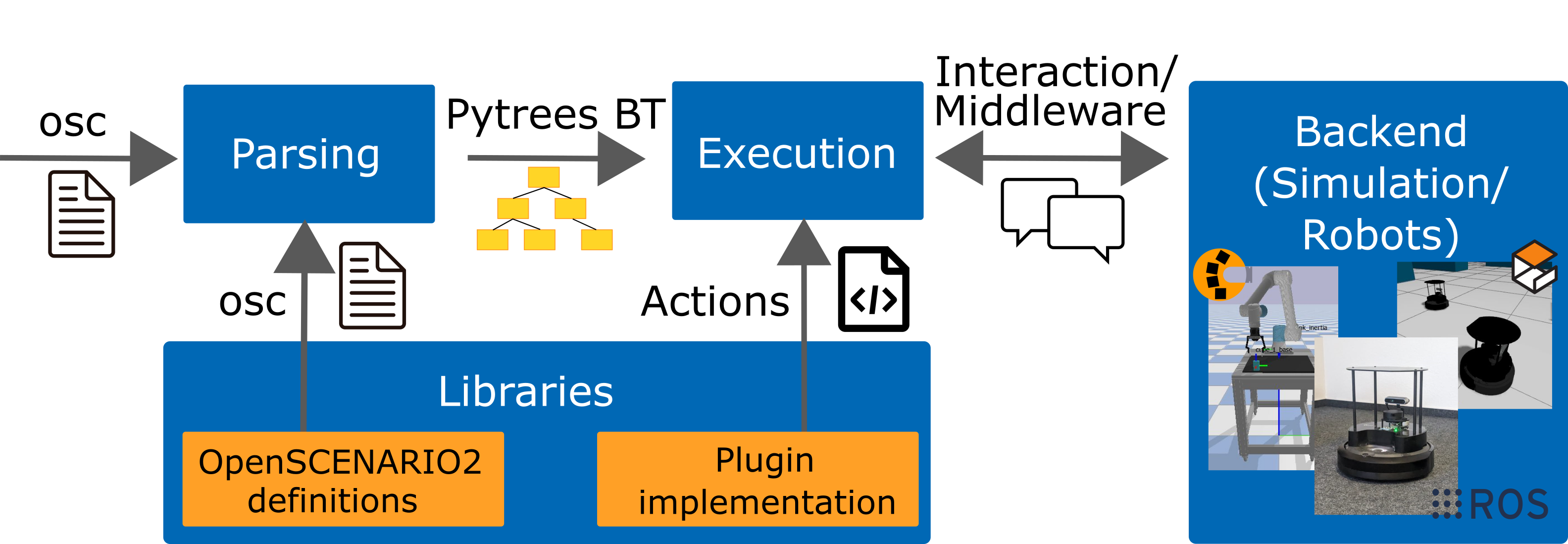}
    \caption{Overview of the modular software architecture of the \acl{SERo} library.}
    \label{fig:software_arch}
    \vspace{-0.5cm}
\end{figure}

Fig.~\ref{fig:software_arch} visualizes the software architecture of our proposed \acl{SERo} library.
Our implementation aims at modularity and flexibility intentionally separating the description of the scenario (see Sec.~\ref{subsec:scenario_description}) and its parsing (see Sec.~\ref{subsec:scenario_parsing}) from the middleware- and backend-specific components such as the simulator, physical robot(s) or the middleware like \ac{ROS2} (see Sec.~\ref{subsec:backed_adaption}).
At its core, \acl{SERo} is built on top of the scenario description language \ac{ASAM} OpenSCENARIO~2 \cite{OpenScenario} and the Python implementation of behavior trees, PyTrees \cite{Pytrees}.
Simulation- or application-specific actions and/or conditions are divided into separate libraries and implemented through a flexible plugin-mechanism (see Sec.~\ref{subsec:plugin_arch}).
Note that libraries typically provide at least an OpenSCENARIO~2 file with additional definitions and may provide code implementing additional functionality such as conditions or actions. 

\subsection{Scenario description}
\label{subsec:scenario_description}

One core component of enabling reproducible experiments, is a generic language for describing the underlying scenarios.
We build our scenario execution library on top of the \ac{ASAM} OpenSCENARIO~2 scenario description language \cite{OpenScenario}, which is a powerful declarative language.
Despite being mainly targeted at automotive scenario testing, OpenSCENARIO~2 does not enforce any standards or API to simulators and testing platforms, which makes it a suitable candidate as front-end for \acl{SERo}. It offers a large feature set, that, e.g., allows parallel and serial execution, the usage of variables and the definition of composite types like actors and actions. Actions form the base of scenario definition and are mapped to actually executed Python implementations through our plugin mechanism.
Listing~\ref{list:simple_nav_scenario} shows a simple example of a scenario description in OpenSCENARIO~2 for an \ac{AMR} navigation use-case.

\begin{figure}[t]
    \centering
    \includegraphics[width=1.\linewidth]{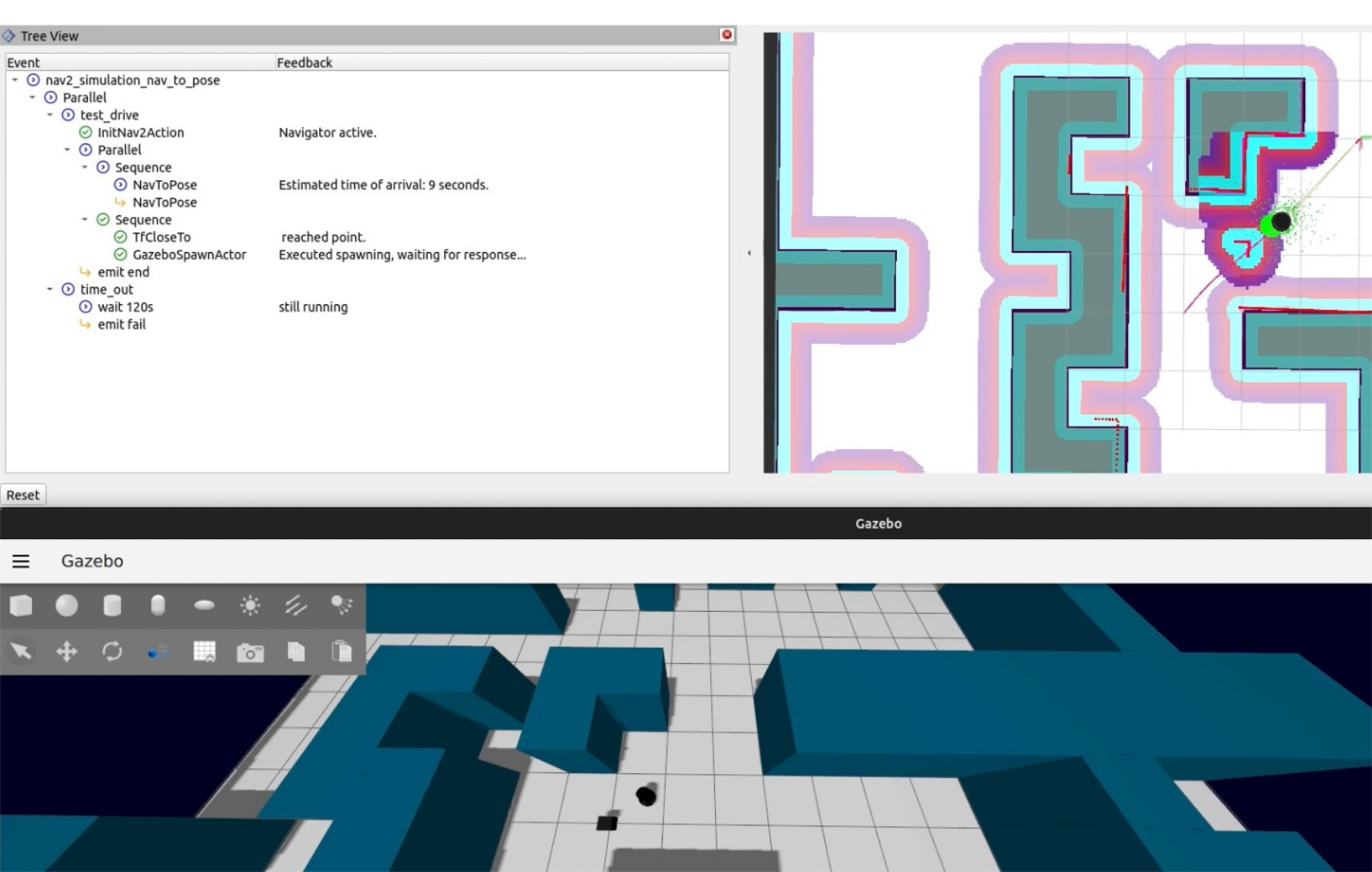}
    \caption{Snapshot of scenario execution realizing a simulated navigation scenario with the Turtlebot4 simulation in Gazebo and \acs{ROS2}.}
    \label{fig:robot_nav_sim_demo}
\end{figure}


\begin{lstlisting}[language=VRML,basicstyle=\scriptsize, belowskip=-0.8 \baselineskip,  float=t, label={list:simple_nav_scenario}, 
caption={\textbf{Simple example scenario file}}]
import osc.ros
scenario example_nav2:
    robot: differential_drive_robot
    do parallel:
        serial:
            robot.init_nav2(pose_3d(position_3d(x: 0.0m, y: 0.0m)))
            robot.nav_to_pose(pose_3d(position_3d(x: 3.0m, y: -3.0m)))
            emit end
        time_out: serial:
            wait elapsed(60s)
            emit fail
\end{lstlisting}

\subsection{Scenario Parsing}
\label{subsec:scenario_parsing}

In this section, we describe the process of parsing a scenario description (osc) file and translating it to a Python behavior tree that can be used by the PyTrees library \cite{Pytrees}.
An \acs{ANTLR4} (\acl{ANTLR4}) \cite{Parr2013} generated parser is used to tokenize the content of the osc file. The resulting token tree is then used as the basis of an internal model, that verifies the consistency of the definitions and that only supported language-features are included. In a next step, all references are resolved and it is checked that the scenario is concrete, i.e., that each parameter has an explicit value (see also Sec.~\ref{subsec:abstract_scenarios}).
Finally, our PyTrees converter translates the internal model to a Python behavior tree using the pyTrees \cite{Pytrees} syntax, behaviors and its blackboard for events.
The final result is a behavior tree in Python, that can be executed either by the PyTrees \cite{Pytrees} library or by its \ac{ROS} counterpart PyTrees ROS \cite{Pytrees-ROS}.

\subsection{Scenario Parameter Variation}
\label{subsec:abstract_scenarios}

One important aspect of scenario-based testing is running the same experiments with different parameters. 
OpenSCENARIO~2 supports this feature by defining lists of values instead of single values. However, as mentioned in Sec.~\ref{subsec:scenario_parsing}, such a scenario is not directly executable.
Hence, our library also includes a tool to translate such a scenario into multiple concrete, executable scenarios, namely one for each permutation of possible parameter values.
For instance, for a scenario with \num{2} varied parameters with \num{8} provided values for each parameter (see Sec.\ref{ssec:varying_params_and_faults} and  Listing~\ref{list:fault_injection_scenario}), this process creates $8^{2}=64$ concrete, executable scenarios.

\begin{figure}[t]
    \centering
    \includegraphics[width=1.\linewidth]{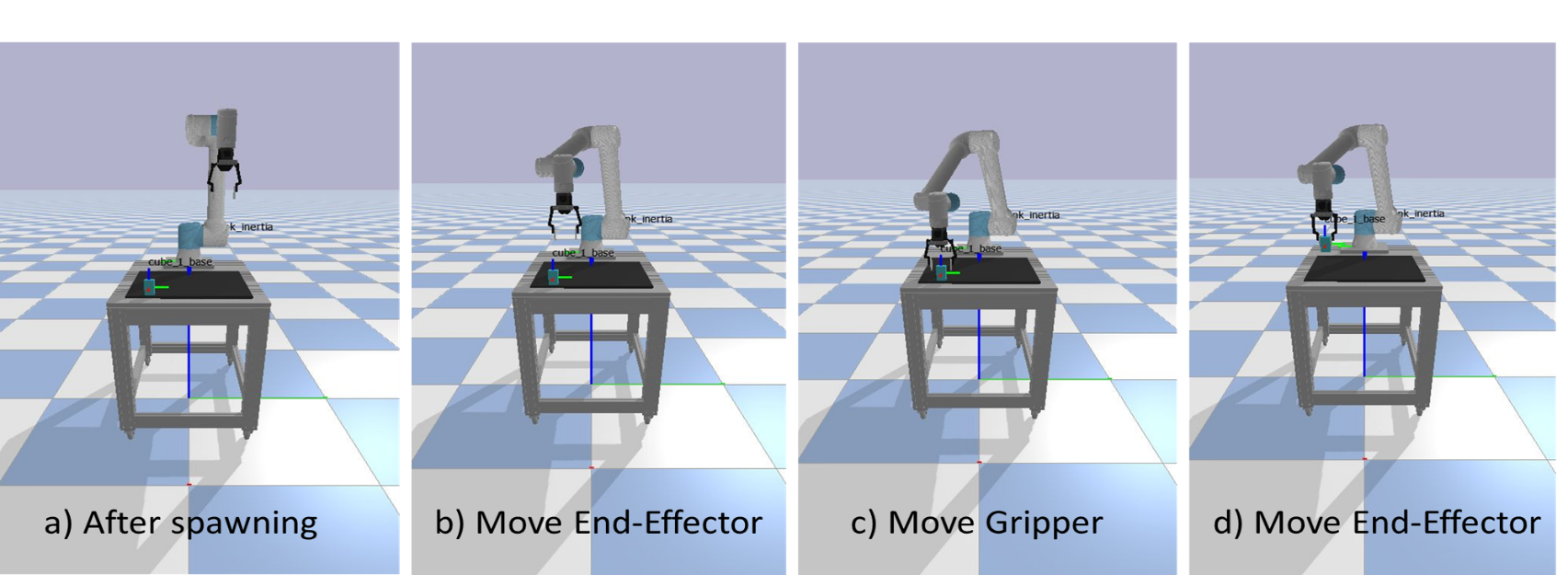}
    \caption{Several selected snapshots of a pick-and-place scenario in the PyBullet simulator realized with \acl{SERo}. From left to right: a) after spawning all actors and objects, b) while moving the end-effector to the grasping position, c) reached grasping position, now move (i.e., close) gripper, d) move end-effector up to reach successful final pick-up pose.}
    \label{fig:pybullet_demo}
    \vspace{-0.5cm}
\end{figure}

\subsection{Middleware/Backend adaptation}
\label{subsec:backed_adaption}

All components of \acl{SERo} described so far, are entirely agnostic of the backend (i.e., specific simulators or physical robots) and the middleware-implementation (e.g., \ac{ROS2}).
As noted in Sec.~\ref{subsec:software_arch}, the backend-specific code to, e.g., spawn and remove objects such as actors or obstacles or to control other aspects of the simulation, is located in libraries separated from the code for parsing and executing the scenarios.
For a specific middleware such as \ac{ROS2}, the execution components of the library make use of the \ac{ROS2}-specific implementation of PyTrees \cite{Pytrees-ROS}.
Additionally, \acl{SERo} already provides the most common \ac{ROS} mechanisms such as publishing data to a \ac{ROS} topic, calling \ac{ROS} services, waiting for data to be published on a specific \ac{ROS} topic.
For a downstream analysis of executed scenarios, \acl{SERo} also supports recording \ac{ROS} bagfiles by a dedicated action, which specifies the list of topics to be recorded to the file.
Thereby, \acl{SERo} will not start with the execution of the actual scenario before all topics to be recorded to the bagfile are available.
Finally, to visualize the scenario and make its status comprehensible to the user, we also provide a plugin for \ac{ROS2}'s visualization GUI RViz\footnote{\url{https://github.com/ros2/rviz}} (see Fig.~\ref{fig:robot_nav_sim_demo}).

\subsection{Plugin-architecture}
\label{subsec:plugin_arch}

Apart from the actual definition and execution of the scenarios, as well as the simulation- and middleware-specific aspects, there are other important ingredients to scenarios, that we realize as plugin libraries for \acl{SERo}.
In principle, any additional feature that is required by a specific scenario and that can be implemented in Python could be realized as additional library.
A library typically provides an OpenSCENARIO~2 file with additional definitions and may provide code implementing additional functionality such as conditions or actions.
One example could be an action that allows \acl{SERo} to control the doors in simulation worlds created using the \ac{Open-RMF}.
Libraries for basic \acs{ROS2} functionality (e.g.,  topic handling, service calls) and Gazebo (e.g., actor spawning) are already available.
We provide tutorials and examples with our code on how to implement custom plugins\footnote{\url{https://intellabs.github.io/scenario_execution}}.

\section{EXPERIMENTS}
\label{sec:experiments}
\begin{figure}[t]
    \centering
    \includegraphics[width=1.\linewidth]{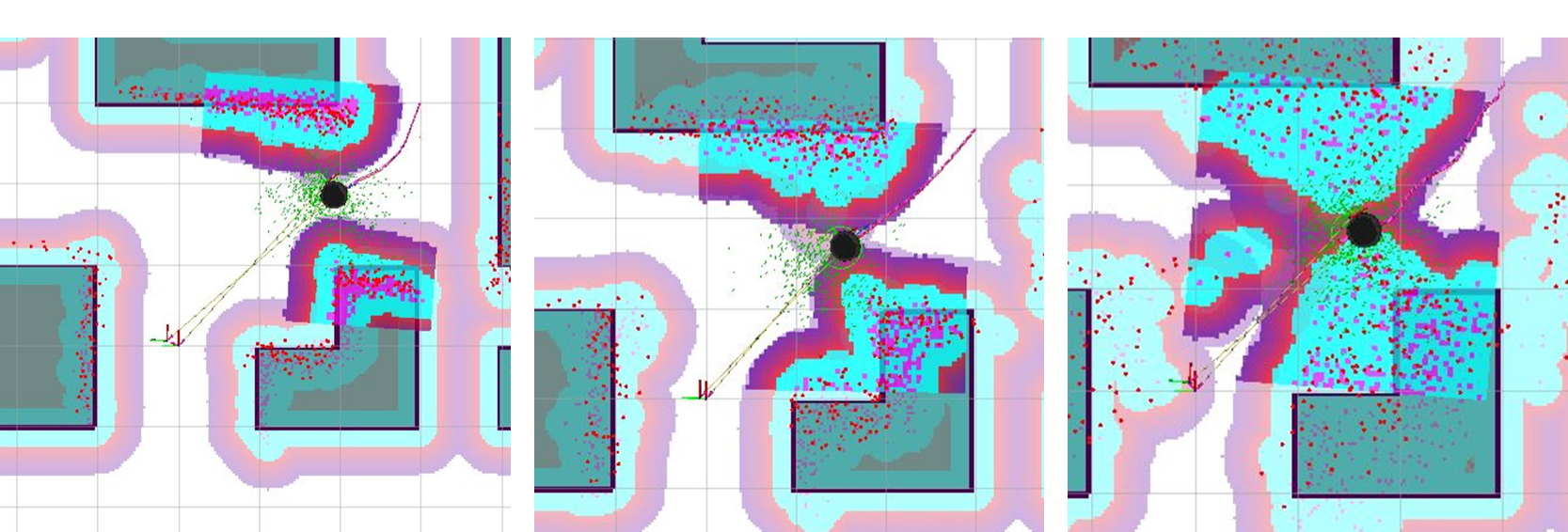}
    \caption{Snapshot of selected runs of the abstract parameter variation scenario described in Sec.~\ref{ssec:varying_params_and_faults} with different values for the standard deviation (namely \num{0.1}, \num{0.2}, \num{0.5} from left to right) of the Gaussian noise added to the LIDAR scan}
    \label{fig:sim_fault_inject}
    \vspace{-0.3cm}
\end{figure}

To demonstrate the feature set of \acl{SERo}, we conduct three different experiments. 
With the first experiment (Sec.~\ref{ssec:simulator_independence}), we demonstrate that \acl{SERo} is independent of the simulation backend and the robotic application by realizing two scenarios in different application domains and different simlators: one \ac{AMR} navigation scenario using \ac{ROS2}'s navigation framework Nav2 \cite{Macenski2020} in Gazebo \cite{Koenig2004} and one manipulation scenario with a robotic gripper in the PyBullet simulator\cite{Zeng2020, PybulletOnline} and a \ac{gRPC} backend that uses \acl{SERo} as pure Python package entirely without \ac{ROS2}.

\subsection{Simulator- and application-independence}
\label{ssec:simulator_independence}

\subsubsection{\ac{AMR} navigation in simulation with Gazebo and \ac{ROS2}}
In this example scenario, we simulate a Turtlebot4\footnote{\url{https://turtlebot.github.io/turtlebot4-user-manual/software/turtlebot4_simulator.html}} in Gazebo and make it navigate to a goal pose in the maze world using Nav2 \cite{Macenski2020} and \ac{ROS2}.
Listing~\ref{list:tb4_scenario} shows the corresponding scenario description file.
\begin{figure}[t]
    \centering
    \includegraphics[width=1.\linewidth]{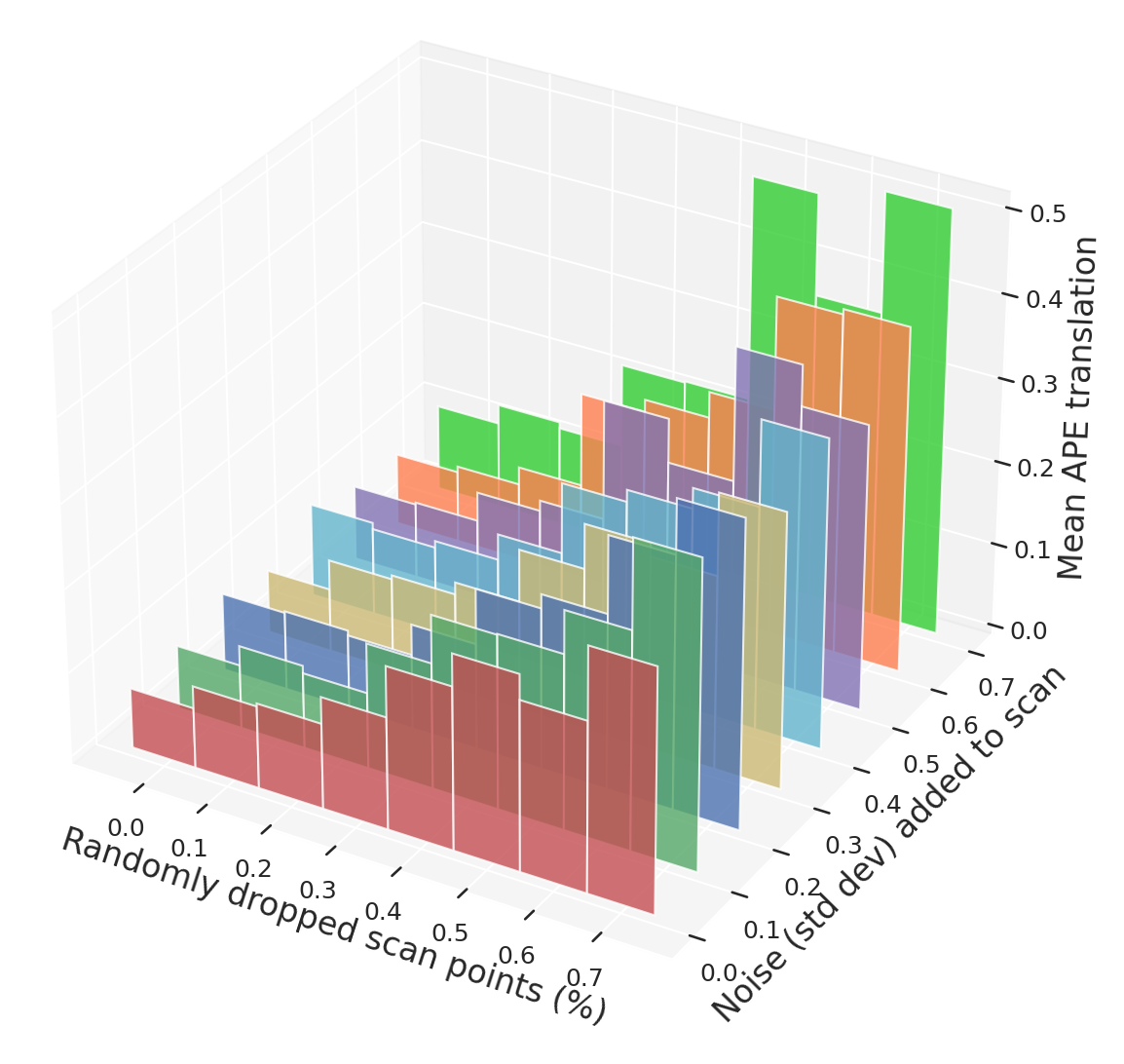}
    \caption{Mean localization error for varying levels of sensor faults injected to the LIDAR scan}
    \label{fig:localization_error_abstract}
\end{figure}

\begin{lstlisting}[language=VRML, belowskip=-0.8 \baselineskip, basicstyle=\scriptsize,  float=t, label={list:tb4_scenario}, 
caption={\textbf{AMR/Nav2 scenario file}: Navigate to a goal pose, spawn an unknown obstacle in the robot's way }]
import osc.ros
import osc.gazebo

scenario example_simulation:
    robot: differential_drive_robot
    box: sim_object
    do parallel:
        serial:
            robot.init_nav2(pose_3d(position_3d(x: 0.0m, y: 0.0m)))
            parallel:
                robot.nav_to_pose(pose_3d(position_3d(x: 3.0m, y: -3.0m)))
                serial:
                    robot.tf_close_to(
                        reference_point: position_3d(x: 1.5m, y: -1.5m),
                        threshold: 0.4m,
                        robot_frame_id: 'turtlebot4_base_link_gt')
                    box.spawn(
                        spawn_pose: pose_3d(
                            position: position_3d(x: 2.0m, y: -2.0m, z: 0.1m),
                            orientation: orientation_3d(yaw: 0.0rad)),
                        model: 'example_simulation://models/box.sdf')
            emit end
        time_out: serial:
            wait elapsed(120s)
            emit fail
\end{lstlisting}
First, the actors (Turtlebot4) and objects (Box) involved in the scenario are defined.
After initializing Nav2, i.e., providing its localization module \ac{AMCL} with the initial pose, the robot is instructed to navigate to its goal position.
In parallel, the \enquote{tf-close-to} action monitors the robot's ground truth position as provided by Gazebo and once the robot is close to the user-specified reference point, a box is spawned right in front of the robot to test its obstacle-avoidance capabilities as the box is not part of the pre-recorded map.
In this simple scenario, the robot successfully avoids the box and reaches its desired goal.
Fig.~\ref{fig:robot_nav_sim_demo} shows a snapshot of this scenario.
In a possible benchmarking experiment, the distance between the reference point and the spawn position of the box could be varied (see Sec.~\ref{subsec:abstract_scenarios}) to evaluate how short of a distance to an abruptly appearing unknown obstacle the robot's navigation stack is able to handle safely.

\subsubsection{Manipulation in simulation with Pybullet}

\begin{lstlisting}[language=VRML, belowskip=-0.8 \baselineskip, basicstyle=\scriptsize,  float=t, label={list:fault_injection_scenario}, 
caption={\textbf{Scenario snippet for injecting faults (Gaussian noise and random drop) into LIDAR scan sensory data}}]
set_node_parameter(
    node_name: 'laserscan_modification',
    parameter_name: 'gaussian_noise_std_deviation') with:
    keep(it.parameter_value in ['0.0', '0.1', '0.2', '0.3', '0.4', '0.5', '0.6', '0.7'])
set_node_parameter() with:
    keep(it.node_name == 'laserscan_modification')
    keep(it.parameter_name == 'random_drop_percentage')
    keep(it.parameter_value in ['0.0', '0.1', '0.2', '0.3', '0.4', '0.5', '0.6', '0.7'])
\end{lstlisting}

In the second example scenario, we realize a manipulation task with a robotic gripper in the PyBullet simulator.
Here, \acl{SERo} runs as a pure Python package without \ac{ROS2}. The functionality to interact with the manipulation software stack and the simulator through \ac{gRPC} is implemented within a custom library.
\begin{figure*}[t]
    \centering
    \includegraphics[width=1.\linewidth]{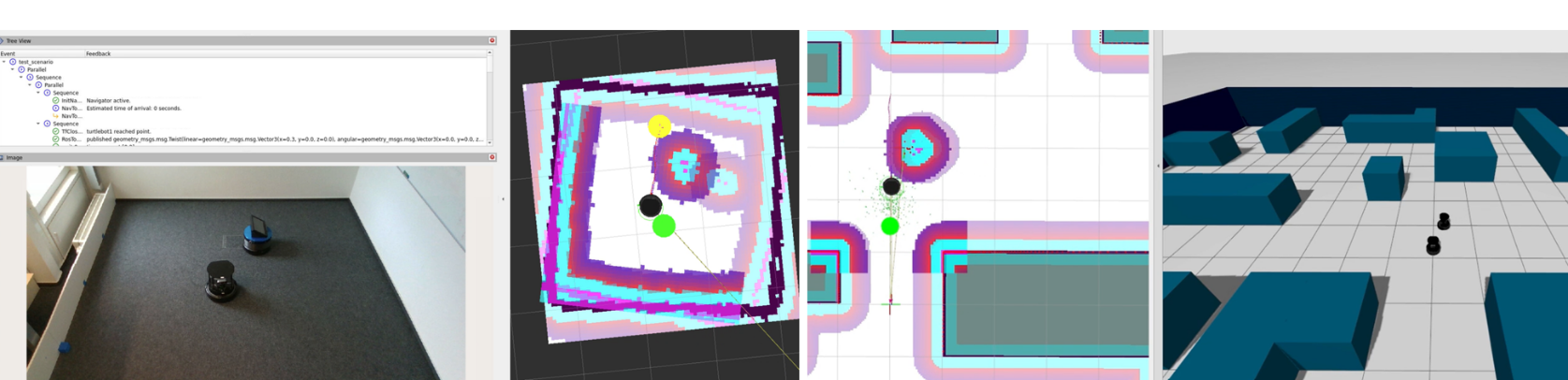}
    \caption{Snapshot of \acl{SERo} realizing a navigation scenario with two physical Turtlebot4 robots (left) and a similar scenario in simulation (right).}
    \label{fig:robot_nav_rw_demo}
    \vspace{-0.5cm}
\end{figure*}
After all actors (\enquote{ur10e-cell} robot) and objects (a camera and the cube object to be picked up) are spawned (Fig.~\ref{fig:pybullet_demo} a), the robot is instructed to move its joints to achieve a suitable position for approaching the cube (\enquote{send-joint-position} action), before the end-effector is moved to the final grasping position (Fig.~\ref{fig:pybullet_demo} b) and the gripper is instructed to close (\enquote{move-gripper} action, Fig.~\ref{fig:pybullet_demo} c).
Finally, the \enquote{move-eef-through-poses} action is called again to make the robot pick up the cube (Fig.~\ref{fig:pybullet_demo} d).
Despite being simple, this scenario demonstrates how \acl{SERo} can be used to evaluate and compare motion planning algorithms for robotic arms in a reproducible setting.
Fig.~\ref{fig:pybullet_demo} shows several snapshots of the scenario.

\subsection{Varying parameters and fault-injection}
\label{ssec:varying_params_and_faults}




In the second use-case, we demonstrate how \acl{SERo} can be used to evaluate a range of varying parameters within a single scenario description file.
Furthermore, this scenario demonstrates how our library enables us to inject faults into the test scenarios to artificially emulate sensor-failures or other safety-critical events in simulation. 
Therefore, we implemented an additional \ac{ROS2} node for fault-injection that manipulates incoming sensor measurements (in our case, 2D-LIDAR scans) by either adding Gaussian noise or randomly dropping a certain percentage of the detections.
The level of injected noise and/or percentage of dropped sensor measurements is controlled via \ac{ROS2} setting the corresponding node parameters from \acl{SERo} over a range of noise/drop percentage values.
We extend the \ac{AMR} scenario shown in Listing~\ref{list:simple_nav_scenario} by the snippet shown in Listing~\ref{list:fault_injection_scenario} to set the failure parameters at the beginning of the scenario.
Fig.~\ref{fig:sim_fault_inject} shows a snapshot of this scenario with Gaussian noise added to the LIDAR scan with varying standard deviation values of \num{0.1}, \num{0.2} and \num{0.5} and no LIDAR points dropped.
Finally, Fig.~\ref{fig:localization_error_abstract} shows the influence of the amount of injected sensor faults on the localization quality of the \ac{AMCL} module becoming worse with an increasing amount of faults injected into the sensory data.
Note that this is one example to demonstrate the parameter variation feature of \acl{SERo} rather than an in-depth localization evaluation.

\subsection{Multi-robot scenarios and sim-to-real transfer}
\label{ssec:real_robot_exp}

In the last example, we demonstrate how \acl{SERo} allows a simple transition from simulation to real-world experiments using  physical robots with minimal adaptations to the scenario file.
Here, we realize an example scenario with two Turtlebot4s: One robot is instructed to navigate to a user-specified goal position using Nav2.
Once this robot reaches a certain reference position, we send a velocity command to the other robot making it drive straight forward into the way of the Nav2-controlled robot.
Similar to the experiment in Sec.~\ref{ssec:simulator_independence}, we use the \enquote{tf-close-to} action to trigger an event, namely the second robot starting to move.
In the simulated experiment, we again use the robot's ground truth position as provided by Gazebo to trigger the event (right image in Fig.~\ref{fig:robot_nav_rw_demo}).
Although there is no ground-truth information available in the real-world experiments, we again employ the \enquote{tf-close-to} action using the robot's position within the map as estimated by its localization module \ac{AMCL} to trigger the start of the second robot's movement (left image in Fig.~\ref{fig:robot_nav_rw_demo}).
Instead of using the robot's self-localization system, we could also employ an external tracking system to generate pseudo-ground truth data of the robot's location.
Fig.\ref{fig:robot_nav_rw_demo} shows a snapshot of both, the simulated and real-world variant of this example scenario at the time when the second robot drives in the Nav2-controlled robots way.
Note that except for the location information (initial pose for \ac{AMCL}, reference point of the \enquote{tf-close-to} action and navigation goal) which depend on the map of the environment and thus differ between simulation and real-world, we use the exact same scenario description file for the simulated and real-world experiment.

\section{DISCUSSION}
\label{sec:discussion}

\subsection{Conclusion}%
\label{subsec:conclusion}

In this paper, we proposed a generic and modular, backend- and middleware-agnostic library for scenario-based, reproducible testing of robotic systems called \acl{SERo}.
Being built on top of the generic scenario description language OpenSCENARIO~2 and Behavior Trees, we make the source code of our library publicly available on GitHub to enable the robotics community to run their experiments in a more systematic, reproducible and automatable way.
We demonstrated, how our approach can be used as stand-alone Python library or in combination with the \ac{ROS2} ecosystem to evaluate robotic systems independent of the simulator or the application domain.
Additionally, we showed how we can use a single scenario definition to execute multiple tests with varying parameters by artificially injecting different faults and thereby test their influence on the overall robotic system.
Finally, we demonstrated that \acl{SERo} also supports the transfer from simulated to real-world experiments with minimal adaptations to the underlying scenario description.

\subsection{Future work}
\label{subsec:future_Work}

Although we believe that \acl{SERo} already supports the most important features of the OpenSCENARIO~2 language for the robotics community, there are still some aspects that are not implemented yet.
Furthermore, although we demonstrated that \acl{SERo} is generally independent of the simulation-backend, we only implemented support for Gazebo and some initial support for PyBullet so far.
With support from the robotics community, we aim to implement backend-specific libraries for other simulators, for instance, \ac{MuJoCo} \cite{Todorov2012}.
Finally, we aim to continue our work on providing a catalogue of relevant scenarios for testing robotics systems starting with the application domain of \aclp{AMR}.
This could lay the foundation for reproducible experiments at scale in the robotics domain when combined with approaches from the cloud-edge-continuum.


\bibliographystyle{IEEEtran} 
\bibliography{IEEEabrv, IROS2024_ScenarioExecutor/refs}

\end{document}